\documentclass[journal]{IEEEtran}

\usepackage{indentfirst}
\usepackage{algorithmic}
\usepackage[linesnumbered,ruled]{algorithm2e}
\usepackage{graphicx}
\usepackage{epstopdf}
\usepackage{amsmath}
\usepackage{cases}
\usepackage{booktabs}
\usepackage[table,xcdraw]{xcolor}
\usepackage{multirow}
\usepackage{array}
\usepackage{color}
\usepackage{enumerate}
\usepackage{cite}
\usepackage{rotating}
\usepackage{pdflscape}
\usepackage{diagbox}

\usepackage{booktabs}
\usepackage{amsmath, amsfonts}
\usepackage{lipsum}
\usepackage{bm}
\usepackage{cleveref}

\usepackage{url}
\usepackage{booktabs}
\usepackage{amsmath, amsfonts}
\usepackage{lipsum}
\usepackage{bm}
\usepackage{amssymb}
\usepackage{threeparttable}
\usepackage{xspace}

\ifCLASSINFOpdf
\else
\fi

\begin{document}
%
\title{Incremental Self-training for Semi-supervised Learning}
%
%

\author{Jifeng Guo,~\IEEEmembership{Member,~IEEE,}
        Zhulin Liu,
        Tong Zhang,~\IEEEmembership{Senior Member,~IEEE,}
        C. L. Philip Chen,~\IEEEmembership{Fellow,~IEEE\\}
\thanks{
This work was funded in part by the National Key Research and Development Program of China under number 2019YFA0706200, in part by the National Natural Science Foundation of China grant under number 62222603, 62076102, and 92267203, in part by the STI2030-Major Projects grant from the Ministry of Science and Technology of the People’s Republic of China under number 2021ZD0200700, in part by the Key-Area Research and Development Program of Guangdong Province under number 2023B0303030001, in part by the Guangdong Natural Science Funds for Distinguished Young Scholar under number 2020B1515020041, and in part by the Program for Guangdong Introducing Innovative and Entrepreneurial Teams (2019ZT08X214). (\emph{Corresponding author: C. L. Philip Chen})}
     \thanks{Jifeng Guo, Zhulin Liu, Tong Zhang and C. L. Philip Chen are with School of Computer Science \& Engineering, South China University of Technology, Guangzhou 510006, China, and are also with Pazhou Lab, Guangzhou 510335, China.
    }
}

%
%


\maketitle

\begin{abstract}
Semi-supervised learning provides a solution to reduce the dependency of machine learning on labeled data. As one of the efficient semi-supervised techniques, self-training (ST) has received increasing attention. Several advancements have emerged to address challenges associated with noisy pseudo-labels. Previous works on self-training acknowledge the importance of unlabeled data but have not delved into their efficient utilization, nor have they paid attention to the problem of high time consumption caused by iterative learning. This paper proposes Incremental Self-training (IST) for semi-supervised learning to fill these gaps. Unlike ST, which processes all data indiscriminately, IST processes data in batches and priority assigns pseudo-labels to unlabeled samples with high certainty. Then, it processes the data around the decision boundary after the model is stabilized, enhancing classifier performance. Our IST is simple yet effective and fits existing self-training-based semi-supervised learning methods. We verify the proposed IST on five datasets and two types of backbone, effectively improving the recognition accuracy and learning speed. Significantly, it outperforms state-of-the-art competitors on three challenging image classification tasks.
\end{abstract}

%
\IEEEpeerreviewmaketitle

\section{Introduction}
\label{sec:intro}
In real-world applications\cite{chen2022novel,liu2016prediction,10076474}, obtaining a substantial amount of labeled data is a time-consuming and labor-intensive task  \cite{hoyer2022daformer,guo2021efficient}. Therefore, effectively utilizing data, especially unlabeled data, has been a longstanding challenge \cite{van2020survey} for machine learning\cite{guo2024dynamic}. Self-training \cite{yarowsky1995unsupervised}, a semi-supervised learning (SSL) technique, has emerged as a promising approach to leverage the abundance of unlabeled data. It involves an iterative process where a model is initially trained on labeled data and then uses its own predictions to generate pseudo-labels for unlabeled instances. These pseudo-labeled data are subsequently incorporated into the training set, and the process is repeated.

Despite its potential, traditional self-training methods face various challenges, including the potential introduction of noisy pseudo-labels, which can lead to model degradation. Several advancements and refinements have been proposed to address these limitations \cite{Wei2021CVPR, Yang2023CVPR}. For example, the Debiased Self-Training(DST) \cite{DST} designs the decoupled classifier heads to avoid the direct effects of incorrect pseudo labels, which achieves state-of-the-art performance using FixMatch \cite{sohn2020fixmatch} and FlexMatch \cite{zhang2021flexmatch} as the backbone.

\begin{figure}[ht]
\vskip 0.2in
\begin{center}
\centerline{\includegraphics[width=\columnwidth]{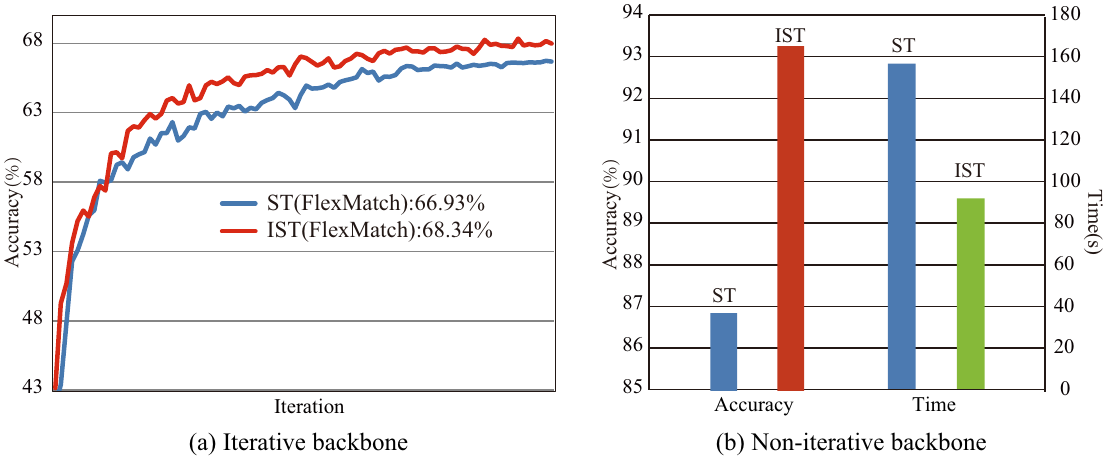}}
\caption{Performance comparison between the ST and IST with two types of backbone. The  iterative backbone is based on the FlexMatch\cite{zhang2021flexmatch} using CIFAR-100, and the non-iterative backbone is based on BLS \cite{Chen2018Broad}.}
\label{fig:comTwo}
\end{center}
\vskip -0.2in
\end{figure}

\noindent \textbf{Motivation:}
Collectively, these previous works demonstrate and emphasize the significance of unlabeled data in improving model performance. However, these works have yet to specifically focus on effectively leveraging unlabeled data or efforts to reduce learning time. Our intuition is that unlabeled data holds immense value in revealing the data distribution, and its rational utilization maybe even more meaningful than training a classifier. Naturally, a question arises: Does leveraging unlabeled data in a progressive manner, from easy to difficult, contribute to improved accuracy, convergence, and reduced learning time?

Aiming to answer this question, we propose Incremental Self-training for semi-supervised learning. The concept of ``incremental" here refers not only to quantity but also to quality. Unlike the undifferentiated utilization in traditional self-training, IST first assigns pseudo labels to positive samples, i.e., unlabeled samples that are easier to classify, to improve the early performance of the base classifier.
In our IST, all unlabeled data are first clustered, and the Euclidean distance between the sample and the cluster center demarcates the certainty of the sample. Based on this certainty, we formed a sequence query list to reduce multiple clustering and repeated queries in the iterative learning process. During the iteration,  we input data in a batch to the base model to improve the model's performance and reduce the learning time or accelerate convergence.
The comparison presented in \cref{fig:comTwo} provides compelling evidence supporting our answer. Our IST framework demonstrates a significant improvement in baseline performance, including accuracy and time. Significantly, it outperforms state-of-the-art competitors on three challenging image classification tasks.

\noindent \textbf{Contribution:} Our contributions can be summarized as:
\begin{itemize}
\item  {Incremental Self-training distinguishes the positivity of unlabeled data by clustering and inputs the model in the sequential batch, which helps improve the base learner's performance. We have tested \textbf{eight} clustering methods to explore the different effects.}
\item {We introduce the sequential query list to reduce the time consumption caused by multiple clustering and queries in iterative learning.}
\item {We implement our IST framework on two types of the backbone of non-iterative and iterative optimization, proving its general suitability. IST achieves an average improvement of 6.41\% compared to standard one and a significant 4\% improvement over state-of-the-art methods on challenging image data sets. In addition, the speed of learning has also improved significantly.}
\end{itemize}

\section{Related Work}
\label{sec:RW}

\subsection{Semi-supervised learning}
Semi-supervised learning encompasses various approaches to leveraging labeled and unlabeled data for improved model performance.
Self-training \cite{oh2022daso,xie2020self} is a popular framework in semi-supervised learning. As a classical self-training approach, Pseudo Label \cite{lee2013pseudo} produces pseudo labels for unlabeled data and updates the model with them continuously. Despite a significant effect, it needs a long training time and may encounter error accumulation problems when learning from incorrect pseudo labels.

Recent studies primarily address this error accumulation in higher-quality and incorrect pseudo labels.
MixMatch \cite{berthelot2019mixmatch} assigns pseudo labels by averaging the results of various augmentations.
Unsupervised Data Augmentation(UDA) \cite{xie2020unsupervised}, ReMixMatch\cite{kurakin2020remixmatch}, and FixMatch \cite{sohn2020fixmatch} employ thresholds of the confidence to control the quality of pseudo labels and forms the strongly augmented samples.
\cite{2021Curriculum} proposed curriculum labeling to improve the quality of pseudo labels during the iteration process. In addition, Dash \cite{Dash} and FlexMatch \cite{zhang2021flexmatch} solve this issue by designing the curriculum learning and dynamical thresholds.
In addition to the density-based consideration, DASO \cite{oh2022daso} also combines it with confidence indicators to generate false labels for each class.

Additionally, the self-supervised learning and adversarial training also solve semi-supervised problems successively.
Self-supervised techniques \cite{he2020momentum} improved the model performance by using the unlabeled data in either the initial pre-training phase \cite{chen2020big} or subsequent downstream tasks \cite{wang2021self}. Nonetheless, the training of self-supervised models typically depends on extensive data and resource-intensive computation, making it impractical for many real-world applications. Adversarial training \cite{ALI} generates fake samples with a ``generated" class label or constructs adversarial samples. For instance, virtual adversarial training(VAT) \cite{miyato2018virtual} introduces additive noise to the input, utilizes adversarial training to estimate the worst-case scenarios related to pseudo labeling, and subsequently avoids such situations. DST \cite{DST} predicted the worst case of self-training bias and adversarially optimized the representations to improve the quality of pseudo labels automatically.

\subsection{Cluster}
Clustering is an unsupervised learning algorithm that can learn underlying relationships from data features and divide the data into clusters. K-means is a fundamental and common method \cite{lloyd1982least}. It finds the centroids of clusters and assigns each data point to the nearest centroid. It has the characteristics of simplicity, scalability, and efficiency. However, it is sensitive to the number of clusters, initial centroid selection, and outliers.

To overcome these limitations, researchers have proposed numerous variants. Fuzzy C-Means (FCM) \cite{dunn1973fuzzy} improves clustering effect and convergence speed by dynamically updating fuzzy membership weights. It can handle unclear data and various cluster shapes, enhancing robustness and interpretability. Hierarchical clustering divides or merges clusters layer by layer, constructing a hierarchical structure and facilitating the exploration of clustering structures at different scales. For example, BIRCH \cite{zhang1996birch} reduces dimensionality using CF trees and employs a hierarchical structure to reduce computational complexity, making it suitable for high-dimensional and large-scale data. Density-based clustering methods, e.g., DBSCAN \cite{ester1996density}, discover clusters of arbitrary shapes based on the density, exhibiting good robustness against outliers and noise. As an extension, OPTICS \cite{ankerst1999optics} determines the clustering structure via reachability distance, enabling the handling of clusters with different densities. MeanShift \cite{comaniciu2002mean} automatically determines the number of clusters and adapts to irregularly shaped clusters but with high computational complexity. Affinity Propagation(AP) \cite{frey2007clustering} has unique advantages in handling uncertain cluster numbers and adapting to irregular shapes or noisy data. However, it has higher computational complexity, especially for large-scale datasets. Spectral clustering \cite{von2007tutorial} maps the similarity matrix of data to a low-dimensional space to perform clustering, overcoming the limitations of K-means in terms of data shape and number of clusters but with higher computational complexity. Mini-batch K-means \cite{sculley2010web} accelerates convergence using mini-batch stochastic gradient descent, making it suitable for large datasets.

\section{Methodology}
\label{sec:formatting}
\textbf{Problem Definition:} The dataset consists of two parts: $n_{l}$ labeled data $\mathcal{D}^{l} =\left\{(x_{i}, y_{i}) \right\}_{i=1}^{n_{l}}$ and $n_{u}$ unlabeled data $\mathcal{D}^{u} =\left\{x_{i} \right\}_{i=1}^{n_{u}}$. Noting, the amount of unlabeled data is typically significantly larger than that of labeled data, i.e., $n_{u}\gg n_{l}$. In the image classification task, the ultimate goal is to find a well-performed model $\psi$ that can achieve a satisfactory recognition effect.

The standard ST framework has impressive performance but treats each unlabeled image equally and neglects its inherent positive and difficulty. This approach may negatively impact the training process due to incorrect predictions in challenging examples. Therefore, we propose Incremental Self-training via cluster and query list to distinguish and prioritize reliable unlabeled data.

\begin{figure}[ht]
\vskip 0.2in
\begin{center}
\centerline{\includegraphics[width=\columnwidth]{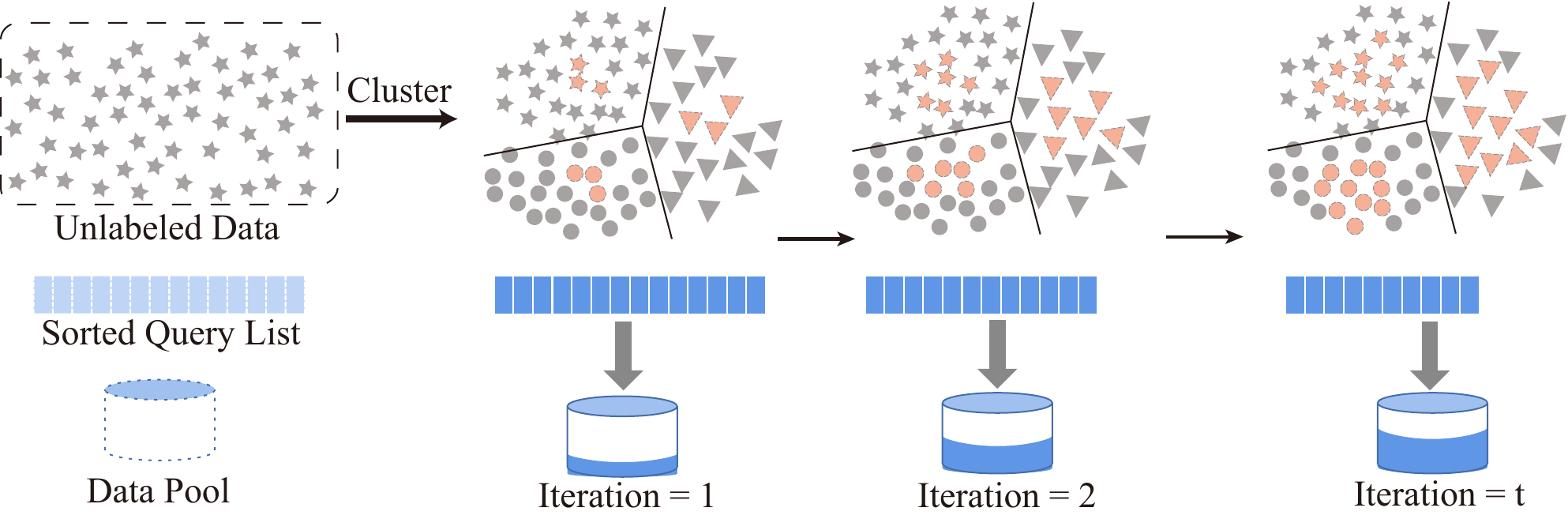}}
\caption{An Illustration of the unlabeled data utilization process. Here, the sample points with color represent being added to the unlabeled data pool for utilization. Cluster distributions are visualizations of query lists to facilitate the understanding of principles.}
\label{fig:cluterPro}
\end{center}
\vskip -0.2in
\end{figure}

Specifically, the performance of the model is enhanced with iteration. Using difficult unlabeled data in the late iteration will effectively improve the model. Instead of evaluating the reliability during iteration \cite{Yang2022CVPR}, our IST distinguishes the difficulty of unlabeled data in the initialization. Considering that the clustering methods can learn the spatial distribution of data in unsupervised way, we adopt clustering to process unlabeled data and further distinguish their positivity.

Similarly, IST consists of three stages: Initialization, Auxiliary Training Data Acquisition, and Classifier Updating. Unlike ST which assigns all unlabeled data in each iteration, unlabeled data is input incrementally and orderly in IST. The clustering methods cluster all unlabeled data during initialization, forming a query list. A fraction of the unlabeled data with the highest certainty, $\textbf{Q}_{a}(0)$, is used to initialize the pool. As the classifier updating progresses, the unlabeled data $\textbf{Q}_{a}(t)$ in the list is sequentially transferred to the pool. 

\section{Experiment}
In the experiments, we employ two types of backbone for validation. We evaluate IST with BLS on two data with different scales, MNIST and EMNIST Digits Dataset. Comparing to the state-of-the-art methods, we evaluate IST with random initialization on standard SSL datasets, including CIFAR-10, CIFAR-100, and SVHN.

For deep backbone training from scratch, Wide ResNet variants\cite{zagoruyko2016wide} are adopted, and its hyperparameters are the same as DST \cite{DST}, including a learning rate of 0.03 and a mini-batch size of 512. We also use the labeled subset consisting of four labels per class to evaluate the performance of IST in scenarios with an acute shortage of labeled data. To ensure a fair comparison, we maintain the consistency of the labeled subset for each dataset across these experiments. Each image undergoes an initial random-resize-crop operation, followed by strong augmentation using RandAugment \cite{Cubuk2020CVPRWorkshops} denoted as $\mathcal{A}$, and weak augmentation with random-horizontal-flip denoted as $a$. In particular, we employ Mini-batch K-means for clustering on CIFAR-100 data to save time.

In confirmatory experiments, we first comprehensively compare our IST and classical ST to assess its effectiveness. All methods are subjected to identical experimental settings to ensure fairness. Without special instructions, we adopt K-Means for clustering in all experiments, and the number of clusters equals the number of categories corresponding to the data set.

\subsection{Ablation studies with clustering methods}
\label{sec:exp1}
We analyze our design as outlined in \cref{tab:Ablation} and \cref{tab:Ablation1}, and identify the following discoveries: (1) Compared with standard self-training, where data are processed without distinction, the sequential processing in incremental self-training where the cluster is responsible for differentiating sample positivity and the sequential query list provide the candidate can better improve the recognition accuracy and learning speed. (2) The effect of different clustering methods is different because that clustering methods have their adaptive data, e.g., MeanShift is suitable for large data and will produce different clustering effects on the same data. (3) Correct recognition of samples at decision boundaries helps improve performance.
\begin{table} [h]
\centering
\caption{Ablation study on iterative backbone.}
\begin{tabular}{l|c|ll}
\toprule
 Method &Cluster\&List   &Acc.(\%) &Time(s)\\
\midrule
ST               &                 &89.30  &57321.65 \\
IST    & w/ K-Means   &93.17\color{red}$\uparrow $ &44796.71
\color{green}$\downarrow$ \\
IST  & w/ MiniBMean   &93.76\color{red}$\uparrow $ &44076.97\color{green}$\downarrow$ \\
IST   &w/ Meanshift   &94.25\color{red}$\uparrow$  &157669.75\color{red}$\uparrow$ \\
 \bottomrule
  \end{tabular}
  \begin{tablenotes}
        \footnotesize
        \item[] The cluster time of Meanshift is 102284.93 seconds.
      \end{tablenotes}
    \label{tab:Ablation}
\end{table}

\begin{table} [h]
\centering
\caption{Ablation study on non-iterative backbone.}
\begin{tabular}{l|c|ll}
\toprule
 Method &Cluster\&List   &Acc.(\%) &Time(s)\\
\midrule
ST               &                 &86.87  &156.63 \\
IST    &w/ BIRCH   &88.97\color{red}$\uparrow$  &99.40\color{green}$\downarrow$ \\
IST     &w/ K-Means   &90.60\color{red}$\uparrow $ &97.85\color{green}$\downarrow$ \\
IST   &w/ MeanShift   &93.28\color{red}$\uparrow $ &91.94\color{green}$\downarrow$ \\
 \bottomrule
  \end{tabular}
    \label{tab:Ablation1}
\end{table}

\begin{figure}[ht]
\vskip 0.2in
\begin{center}
\centerline{\includegraphics[width=0.97\columnwidth]{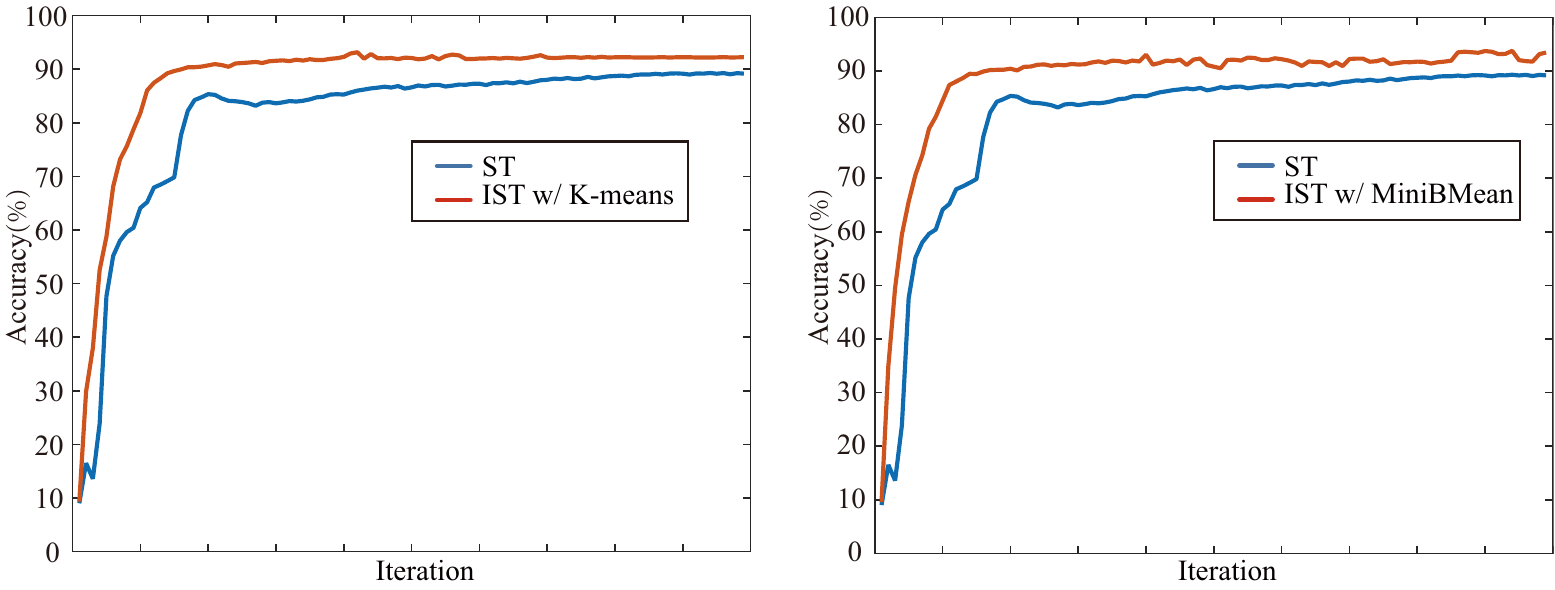}}
\caption{Comparison of ST and IST with iterative backbone.}
\label{fig:cifarTwo}
\end{center}
\vskip -0.2in
\end{figure}

\begin{figure}[ht]
\vskip 0.2in
\begin{center}
\centerline{\includegraphics[width=\columnwidth]{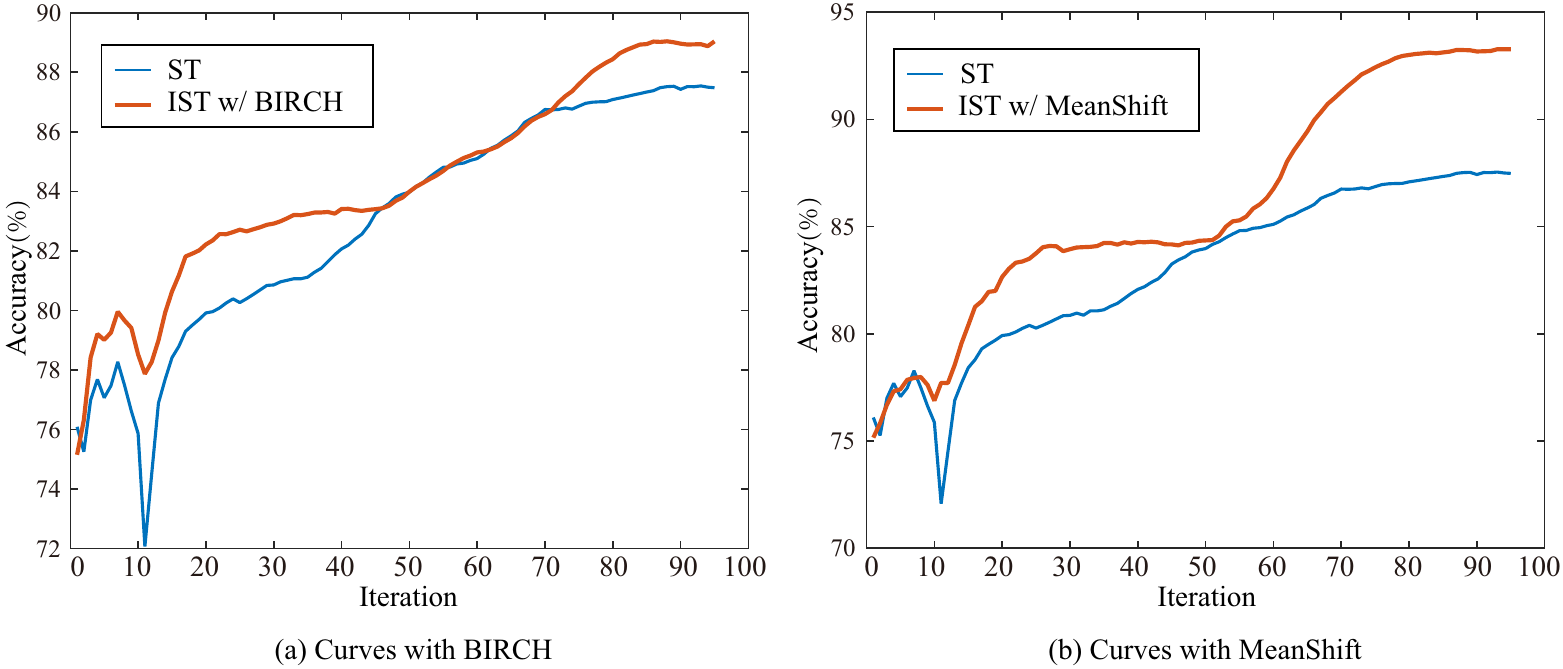}}
\caption{Comparison of ST and IST with non-iterative backbone.}
\label{fig:pro1}
\end{center}
\vskip -0.2in
\end{figure}

\cref{fig:cifarTwo} and \cref{fig:pro1} provide qualitative comparisons, where IST effectively improves the accuracy and convergence speed of the iterative backbone. In addition, it also avoid some accuracy losses that occur in ST with non-iterative backbone. Taking the presented curves in \cref{fig:pro1}.a) as an instance, ST suffered from significant errors in the 11th iteration, dropping from an accuracy of 76\% to 72\%, while IST successfully reduced this loss. Furthermore, our IST can effectively distinguish the unlabeled data around the decision boundary and improve the recognition accuracy.


\begin{figure*}[t]
  \centering
   \includegraphics[width=0.91\linewidth]{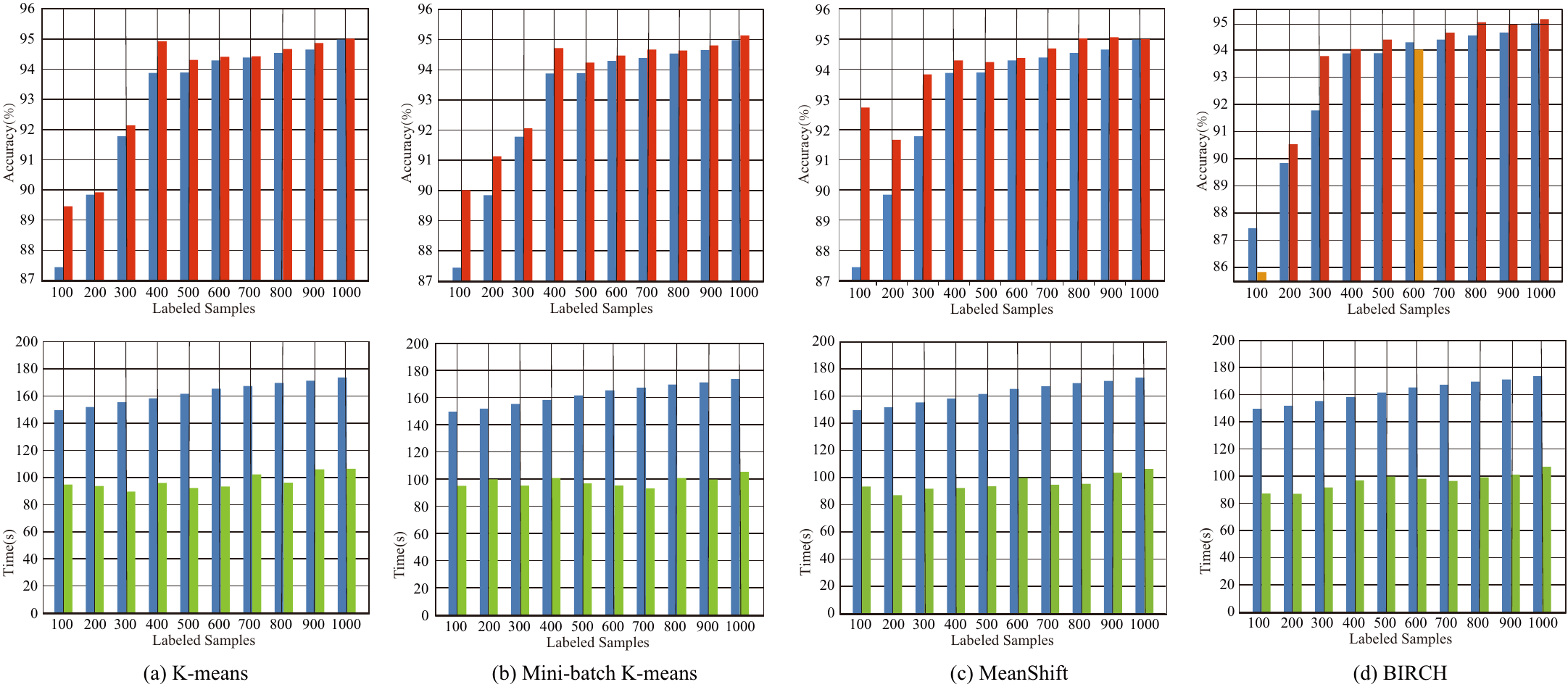}
   \caption{Comparison of accuracy and time in different clustering methods and data. In the figure, orange represents a decrease in accuracy.}
   \label{fig:cluter}
\end{figure*}

\subsection{Performance under different data scales}
\label{sec:exp2.2}

We further derive IST, which utilizes different clusters to promote exploration. \cref{fig:cluter} compares performance with four different clustering methods. It is evident that IST outperforms ST across various numbers of labeled data samples. Specially, IST with MeanShift improves accuracy from 87.44\% to 92.73\%. It is undeniable that the quality of clustering results can impact performance improvements. For instance, the accuracy of IST with BIRCH decreases by up to 1.6\% for 100 and 600 samples.

\begin{figure}[h]
  \centering
   \includegraphics[width=0.91\linewidth]{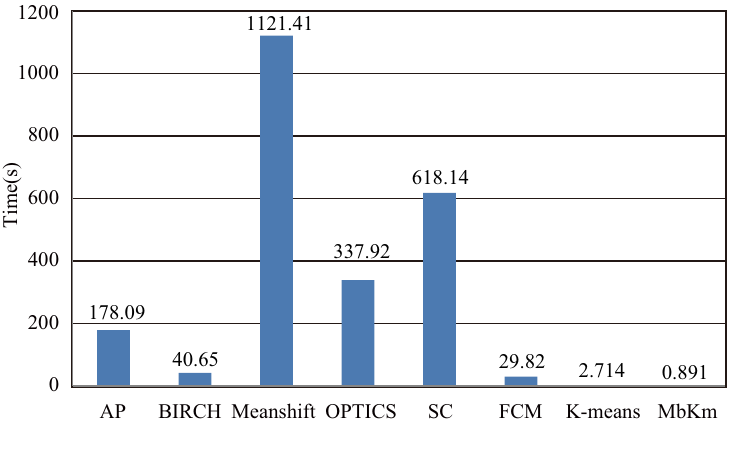}
   \caption{Average clustering time of different method. SC and MbKm stand for spectral clustering and Mini-batch K-means, respectively.}
   \label{fig:time}
\end{figure}

Furthermore, IST significantly reduces learning time consumption. As depicted in \cref{fig:cluter}, the effect of different methods is approximate, and ST's time consumption is reduced by approximately 40\%-50\% across all clusters. However, there are notable differences in clustering time among the various methods. As illustrated in \cref{fig:time}, K-means and Mini-batch K-means exhibit the lowest time consumption, yielding satisfactory results on common data. Conversely, MeanShift has the longest clustering time, up to 1121.41 seconds. Therefore, it is only recommended if irregular clusters exist.


\section{Conclusion}
\label{sec:RW}
In this paper, we introduce a promising framework for semi-supervised learning, Incremental Self-training, addressing the limitations of traditional self-training methods. As an improved variant to classical self-training, our IST processes unlabeled data discriminately. This technique uses clustering methods to distinguish the positivity of data and introduces a sorted query list built according to activity to reduce multiple clustering and query interactions. This work is the first time improving self-training methods regarding recognition accuracy and training time. The experimental results show that the proposed IST method can adapt to various popular models and improve accuracy and training time.

\textbf{Discussion:} The clustering effect affects the performance. We introduce the clustering method to process unlabeled data. According to the report, IST improves the model's performance. It is undeniable that there will be some alarming results when the clustering effect is unsatisfactory. Therefore, choosing the appropriate clustering method according to the data is critical. For example, some clusters are suitable for spherical clustering, while others are suitable for more complex shapes, such as non-convex clustering. We will further explore performance on more data and tasks.

\bibliographystyle{IEEEtran}
\bibliography{main}

\ifCLASSOPTIONcaptionsoff
  \newpage
\fi

\end{document}